%% file: main.tex
\documentclass[10pt,twocolumn,letterpaper]{article}

 \usepackage{cvpr}              


\definecolor{cvprblue}{rgb}{0.21,0.49,0.74}
\usepackage[pagebackref,breaklinks,colorlinks,allcolors=cvprblue]{hyperref}
\usepackage{algorithm}
\usepackage{algpseudocode}
\usepackage{float}

\def\paperID{*****} 
\def\confName{CVPR}
\def\confYear{2025}

\title{LoomNet: Enhancing Multi-View Image Generation via Latent Space Weaving}

\author{
Giulio Federico$^{1,2}$ \quad
Fabio Carrara$^{2}$ \quad
Claudio Gennaro$^{2}$ \quad
Giuseppe Amato$^{2}$ \quad
Marco Di Benedetto$^{2}$ \\
$^1$University of Pisa, Italy, \quad
$^2$CNR-ISTI, Pisa, Italy \\
{\tt <name.surname@isti.cnr.it}
}

\begin{document}

\twocolumn[{%
\renewcommand\twocolumn[1][]{#1}%
\maketitle
\includegraphics[width=\linewidth]{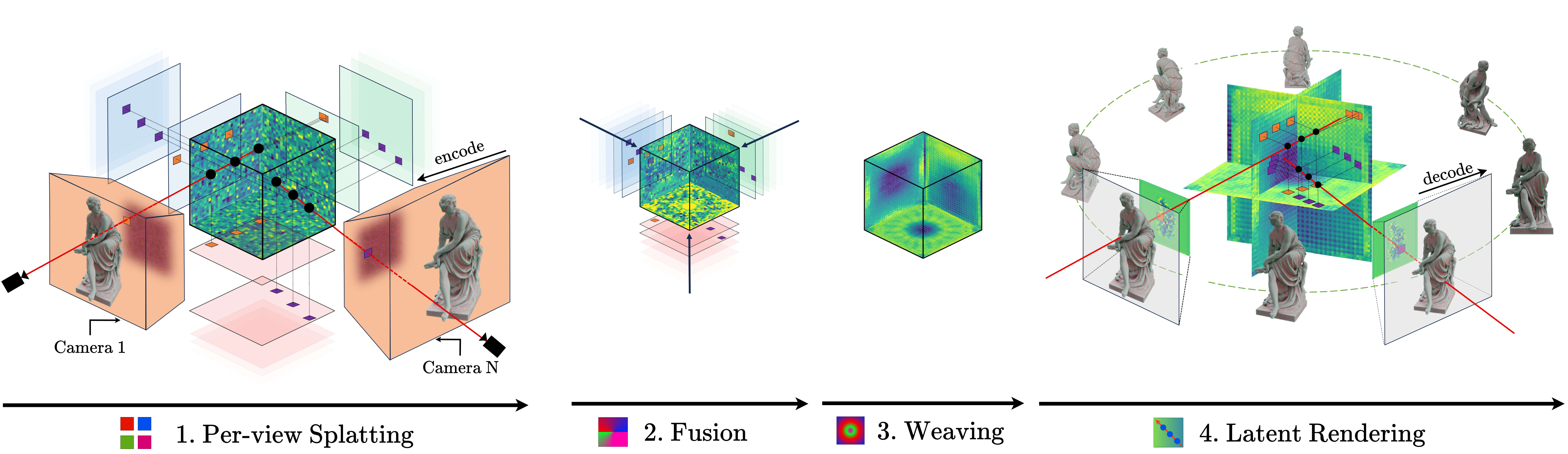}
\captionof{figure}{
Given a set of desired novel perspectives of an input image, \textbf{LoomNet} rapidly generates consistent multi-view images by \textit{rendering} within a \textit{shared latent space}, built by forming per-view hypotheses of the scene (\textit{per-view splatting}), fusing them to reach a discrete multi-view consensus (\textit{fusion stage}), and finally refining it to create a coherent, spatially continuous latent representation (\textit{weaving stage}).}
\label{fig:teaser}
}]

\input{sec/0_abstract}    
\input{sec/1_intro}

\input{sec/2_related_work}

\input{sec/3_method}

\input{sec/4_experiment}
\input{sec/5_limitation_and_conclusion}

{
    \small
    \bibliographystyle{ieeenat_fullname}
    \bibliography{main} 
}


\end{document}

%% file: sec/0_abstract.tex
\begin{abstract}
Generating consistent multi-view images from a single image remains challenging. Lack of spatial consistency often degrades 3D mesh quality in surface reconstruction. To address this, we propose \textbf{LoomNet}, a novel multi-view diffusion architecture that produces coherent images by applying the same diffusion model multiple times in parallel to collaboratively build and leverage a shared latent space for view consistency. Each viewpoint-specific inference generates an encoding representing its own hypothesis of the novel view from a given camera pose, which is projected onto three orthogonal planes. For each plane, encodings from all views are fused into a single aggregated plane. These aggregated planes are then processed to propagate information and interpolate missing regions, combining the hypotheses into a unified, coherent interpretation. The final latent space is then used to render consistent multi-view images. LoomNet generates 16 high-quality and coherent views in just 15 seconds. In our experiments, LoomNet outperforms state-of-the-art methods on both image quality and reconstruction metrics, also showing creativity by producing diverse, plausible novel views from the same input.
\end{abstract}

%% file: sec/1_intro.tex
\section{Introduction}
\label{sec:intro}

Humans, based on prior experience, possess a remarkable ability to perceive 3D structures from a single image. They can easily imagine how an object might appear from various viewpoints and, ultimately, infer its underlying 3D structure. However, transferring this ability to generative algorithms remains a significant challenge. This difficulty arises from two main factors: first, the limited diversity and representativeness of publicly available 3D datasets and second, the inherent complexity of accurately interpreting spatial structure and fine details from a single 2D perspective.
Achieving this capability would profoundly benefit numerous applications, including augmented/virtual reality, robotic simulation and gaming by democratizing and significantly accelerating the creation of 3D models from single images. This would enable systems to perceive the potential volumetric structure of observed scenes, thereby enhancing their understanding of the physical world.
Consequently, the task of generating consistent multi-view images from a single input image remains particularly challenging due to the limited 3D information inherent in a single 2D observation. Even minor inconsistencies among the generated images can severely degrade the quality of subsequent 3D shape reconstruction and appearance (texture) mapping, leading to incomplete geometry or distorted appearance, even when employing advanced rendering and reconstruction systems such as NeRF \cite{mildenhall2021nerf} and NeuS \cite{wang2021neus}.

Recent advances in 3D generation from a single image can be categorized into three main paradigms: (i) direct 3D model generation by training 3D generative models on 3D datasets \citep{anciukevivcius2023renderdiffusion, cao2023large, chen2023single, cheng2023sdfusion, gao2022get3d, gu2024control3diff, gupta20233dgen, jun2023shap, kim2023neuralfield, liu2023meshdiffusion, luo2021diffusion, muller2023diffrf, nichol2022point, ntavelis2023autodecoding, wang2023rodin, wang2023pushing, vahdat2022lion, mo2023dit, ren2024xcube, li2023diffusion, lyu2023controllable}; (ii) distillation of pretrained 2D generative models to supervise 3D optimization \citep{poole2022dreamfusion,chen2023fantasia3d, jain2022zero, melas2023realfusion, qian2023magic123, raj2023dreambooth3d, sun2023dreamcraft3d, tang2023make, wang2023score, wang2023prolificdreamer, yu2023text, yi2024gaussiandreamer}; and (iii) multi-view synthesis from a single input image \citep{liu2023zero, liu2023syncdreamer, huang2024epidiff}, followed by 3D reconstruction \cite{mildenhall2021nerf, wang2021neus}.

While direct generation suffers from limited data availability and distillation methods often require costly optimization, multi-view synthesis has emerged as a promising approach that balances quality and efficiency. However, maintaining consistency across synthesized views and obtaining a unified scene representation remain open challenges. Prior works such as SyncDreamer \cite{liu2023syncdreamer} introduce a global 3D feature volume to aggregate scene information for view consistency but are computationally intensive. More efficient methods like EpiDiff \cite{huang2024epidiff} enforce epipolar constraints to speed up generation but lack an explicit latent scene representation.

In this paper, we focus on multi-view synthesis and propose a method to improve consistency across multiple views of the same scene. Our goal is to create a unified scene representation while keeping generation both high-quality and efficient. We train N diffusion models in parallel, each receiving the same input image but focusing on a different camera pose. Each model encodes its hypothesis of the scene from its viewpoint, projecting it onto three latent planes—a process we call \textit{per-view splatting}. These planes are combined in a \textit{fusion stage} to reach a shared agreement aligning visual and geometric information. Since this fused representation reflects only discrete consensus at sampled views, we introduce a \textit{weaving stage} that refines and connects latent features, effectively “weaving” information across viewpoints into a seamless scene embedding. This process inspires the name of our method, \textbf{LoomNet}. Finally, all views are \textit{rendered} from these shared latent planes, ensuring consistent and coherent multi-view generation (Figure \ref{fig:teaser}).
\newline
Our main contributions are:
\begin{itemize}
\item A new multi-view diffusion architecture that enables efficient communication across views and shared rendering for strong consistency.
\item A unified 3D latent scene representation that supports downstream tasks.
\item State-of-the-art results in both image generation and 3D reconstruction (evaluated with PSNR, SSIM, LPIPS, Chamfer Distance and Volume IoU), with fast inference—producing 16 views in just 15 seconds.
\end{itemize}

%% file: sec/2_related_work.tex
\section{Related work}
\label{sec:related_work}


\subsection{Direct 3D Generation from a Single Image}

The first approach, motivated by the remarkable success of diffusion models \cite{ho2020denoising} for 2D image generation, involves training generative models directly on 3D datasets. A series of research \citep{anciukevivcius2023renderdiffusion, cao2023large, chen2023single, cheng2023sdfusion, gao2022get3d, gu2024control3diff, gupta20233dgen, jun2023shap, kim2023neuralfield, liu2023meshdiffusion, luo2021diffusion, muller2023diffrf, nichol2022point, ntavelis2023autodecoding, wang2023rodin, wang2023pushing, vahdat2022lion, mo2023dit, ren2024xcube, li2023diffusion, lyu2023controllable} have explored training 3D generative diffusion models from scratch on 3D assets, directly generating outputs such as point clouds \citep{mo2023dit, luo2021diffusion, nichol2022point, vahdat2022lion, lyu2023controllable}, meshes \citep{gao2022get3d, liu2023meshdiffusion, ren2024xcube,li2023diffusion}, or even neural radiance fields \citep{wang2023rodin,anciukevivcius2023renderdiffusion,chen2023single,gu2024control3diff,gupta20233dgen,jun2023shap,kim2023neuralfield, muller2023diffrf, erkocc2023hyperdiffusion}. However, despite the continuous growth of public 3D datasets like Objaverse \cite{deitke2023objaverse}, their limited size remains insufficient to capture the diversity and complexity required for generalizable 3D generation, in contrast to the abundance of data available for 2D models. Consequently, these methods often exhibit limited generalizability, producing shapes restricted to specific categories and struggling to adapt to more intricate input conditions.

\subsection{3D Inference via 2D Knowledge Distillation}

Following the introduction of DreamFusion by Poole et al. \cite{poole2022dreamfusion}, a growing number of works \citep{chen2023fantasia3d, jain2022zero, melas2023realfusion, qian2023magic123, raj2023dreambooth3d, sun2023dreamcraft3d, tang2023make, wang2023score, wang2023prolificdreamer, yu2023text, yi2024gaussiandreamer} have explored the distillation of pretrained 2D generative models to supervise the optimization of 3D scene parameters (e.g., NeRF, SDF, mesh). This is typically achieved via Score Distillation Sampling (SDS), which leverages the gradients of diffusion models to guide 3D reconstruction. Despite its effectiveness, this per-view optimization strategy presents several practical limitations. It often requires extensive hyperparameter tuning to mitigate common artifacts—such as duplicate geometry (Janus artifacts) and texture distortions—and tends to be computationally expensive, with optimization taking tens of minutes to hours per scene. Moreover, to fully leverage text-to-image models, these methods frequently rely on text prompt engineering or textual inversion to find suitable conditioning, further increasing complexity.

\subsection{Multi-View Generation for 3D Reconstruction}

Rather than relying on pretrained 2D generative models purely as priors for 3D content creation, recent works \citep{liu2023zero, liu2023syncdreamer, huang2024epidiff} have shifted toward directly fine-tuning such models to synthesize multi-view-consistent images from a single input view. These generated views are subsequently fed into 3D reconstruction pipelines \cite{mildenhall2021nerf, wang2021neus} to recover the object’s geometry. A key challenge in this paradigm lies in ensuring consistency across the synthesized views—a critical requirement for accurate and high-fidelity 3D reconstruction. For example Zero-1-to-3 \cite{liu2023zero} directly produce multiple novel views of an object given a single image prompt but without an explicit mechanism for cross-view communication or structural agreement, because each view is synthesized independently. The generated views often suffer from noticeable inconsistencies: this misalignment not only reduces visual fidelity, but also severely degrades the quality of subsequent 3D reconstruction. Indeed, even advanced reconstruction pipelines like One-2-3-45 \cite{liu2023one} applies a neural reconstruction method for sparse views \cite{long2022sparseneus} to the multi-view images produced by Zero-1-to-3, but the lack of coherent cross-view structure leads to incomplete or geometrically implausible meshes. SyncDreamer \cite{liu2023syncdreamer} addresses this issue by introducing a global 3D feature volume that aggregates and synchronizes visual information across views during generation. However, this approach is computationally heavy, slow and also tends to produce low-quality images. In contrast, EpiDiff \cite{huang2024epidiff} enforces communication between views through epipolar constraints, achieving significantly higher efficiency and quality. Though EpiDiff achieves better quality and consistency, lacking a shared space makes it prone to incoherent error propagation. SyncDreamer’s shared space yields smoother but sometimes semantically wrong surfaces with coherent errors.

%% file: sec/3_method.tex
\section{Method}

Given an input view \(V^0 \in \mathbb{R}^{H \times W \times 3}\) of an object, our goal is to generate \(N\) additional views of the same object from different viewpoints, each defined by a rotation matrix \(R \in \mathbb{R}^{3 \times 3}\) and a translation vector \(T \in \mathbb{R}^3\). Formally, we aim to train a model \(f\) to synthesize these novel views \(V^{1:N}\), conditioned on the input view and the target camera poses

\begin{equation}
V^{1:N} = f\left(V^{0}, [R,T]^{1:N}\right) \,.
\end{equation}

This function is modeled via \(N\) parallel instances of the same Latent Diffusion Model (LDM) \cite{rombach2022high} \(f_\theta^{1:N}\), each responsible for generating one view. In the LDM framework, images are first encoded into a latent space via an encoder \(\mathcal{E}\) and then reconstructed with a decoder \(\mathcal{D}\), enabling diffusion to be performed in latent space instead of RGB image space.

In the multi-view generation setting, each model \(f_\theta^i\) approximates the reverse diffusion process \(q(z_{t-1}^i \mid z_t^i, z^0, [R,T]^i)\). When learned correctly, starting from pure noise \(z_T^i\), the model denoises through \(\mathcal{T}\) steps to recover a clean latent \(z_0^i\). Generation is conditioned on the reference latent \(z^0 = \mathcal{E}(V^0)\) and the relative viewpoint transformation \([R,T]^i\) with respect to the input view \([R,T]^0\). Decoding \(z_0^i\) via \(\mathcal{D}\) yields the synthesized view \(V^i\).

For example, Zero1-to-3 \cite{liu2023zero} trains the model by minimizing:
\begin{equation}
    \mathcal{L}(\theta) = 
    \mathbb{E}_{z^0, t, \varepsilon} 
    \left[
        \left\| \varepsilon - f_\theta^i(z_t^i, z^0, t, [R,T]^i) \right\|_2^2
    \right] \,,
\label{eq:loss_zero123}
\end{equation}
where at each diffusion timestep \(t=1,\dots,\mathcal{T}\), the model predicts a progressively denoised latent \(z_t^i\) approaching \(z_0^i\). Each \(f_\theta^i\) is a UNet composed of 4 encoder blocks, a bottleneck and 4 decoder blocks, trained to remove noise from the input latent \(z_t^i\).

\begin{figure*}[t]
    \centering
    \includegraphics[width=\textwidth]{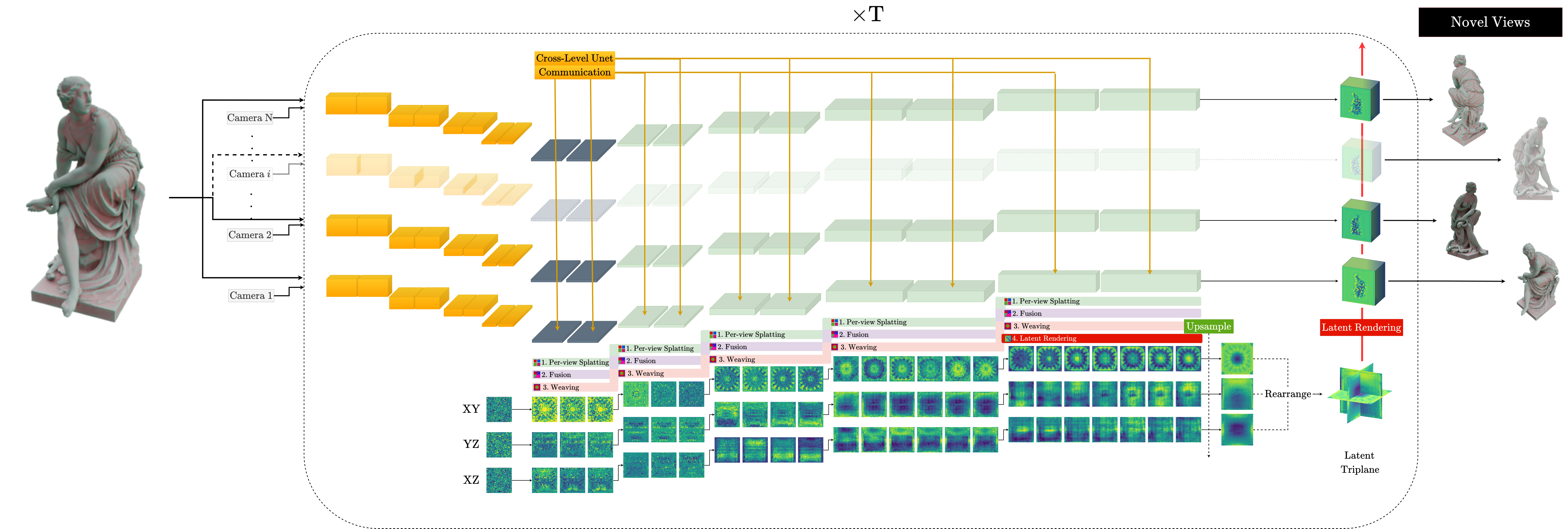}
    \caption{Architecture overview. Given an input image and target viewpoints, each model independently encodes its view via a shared UNet. At the bottleneck and decoder, the \textit{splatting, fusion and weaving stages} aggregate information into a shared latent space, from which latent rendering produces the final views.}
    \label{fig:architecture_overview}
\end{figure*}

\textbf{To generate multiple consistent views} of the same object, it is crucial that the models producing each view communicate during denoising. \textit{Consistent and coordinated synthesis can only be achieved if the noise removal process in each view is aware of the other views’ progress.} To this end, we introduce in each decoder block a \textbf{communication module} among the \(N\) UNets, aggregating information on how each model removes noise from its perspective. This module enables each UNet to adjust its output in a coordinated manner, accounting for others’ strategies to achieve the desired consistency. The module is integrated at each of the 4 decoder levels and consists of \textbf{four stages}: \textit{1) per-view splatting}, \textit{2) fusion}, \textit{3) weaving} and \textit{4) latent rendering} (the latter only in the final decoder level, see Figure~\ref{fig:architecture_overview}).

\subsection{Per-View Splatting Stage}

In this initial phase, each model produces, at a given diffusion timestep, its own hypothesis of the scene from its assigned viewpoint. This hypothesis is represented as a feature map \( F \in \mathbb{R}^{H' \times W' \times C} \), which may be more or less coarse depending on the decoder depth.

The key idea is that if multiple cameras observe the same 3D point, \textit{that point can be used to aggregate pixel features} from the corresponding feature maps of all observing views, enabling coherent synchronization during training.

\begin{figure}[H]
    \centering
    \includegraphics[width=0.92\columnwidth]{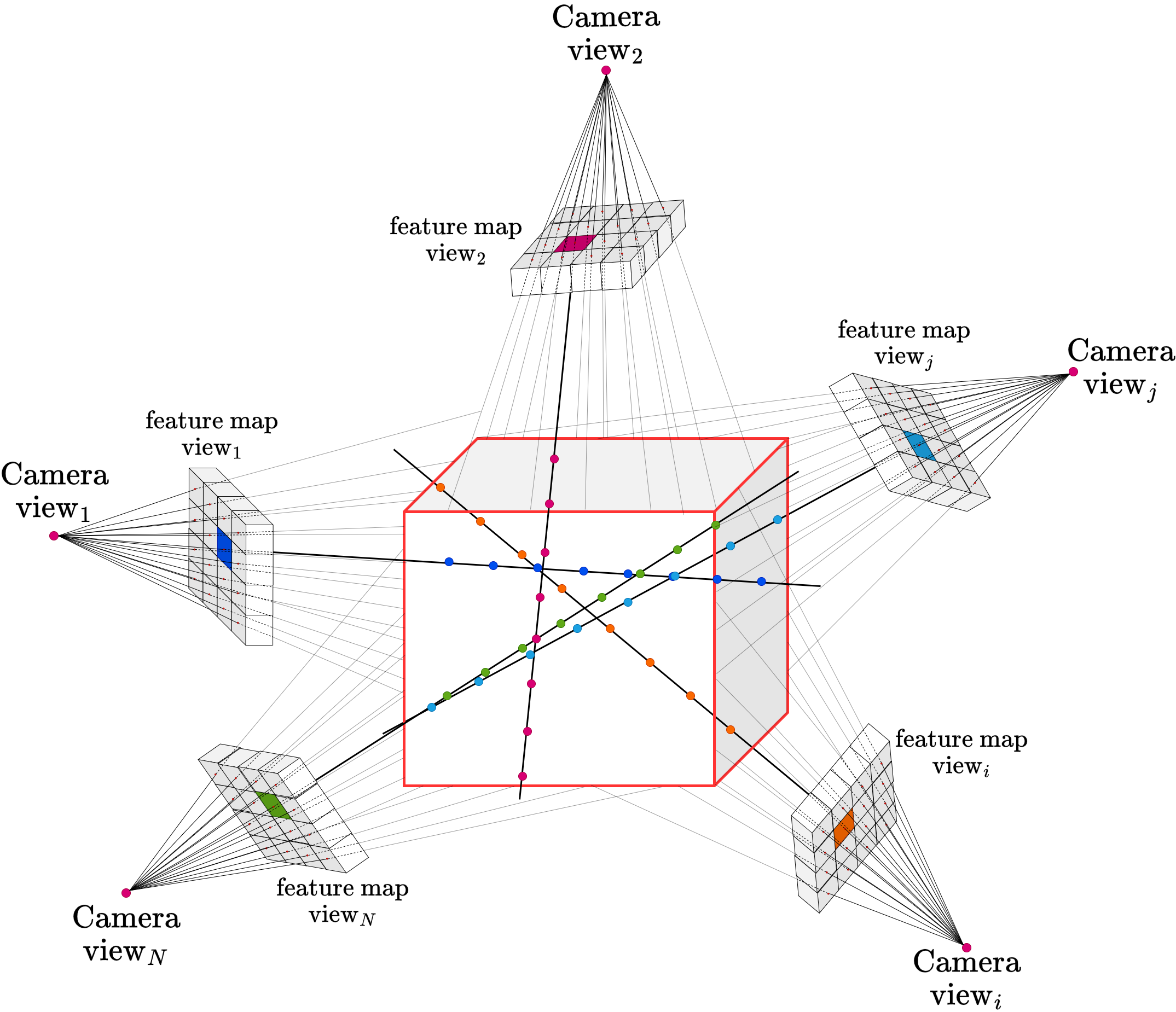}
    \caption{Extraction of rays from each model’s feature map for view sharing within a cube centered at the origin.}
    \label{fig:cube_splatting}
\end{figure}

To share its hypothesis with other models, each feature map \(F\) generates \( H' \times W' \) rays: each ray originates from the camera center and propagates through the direction of a pixel in \(F\), intersecting a cube centered at the scene origin both on entry and exit (Figure \ref{fig:cube_splatting}).

\begin{figure}
    \centering
    \includegraphics[width=\columnwidth]{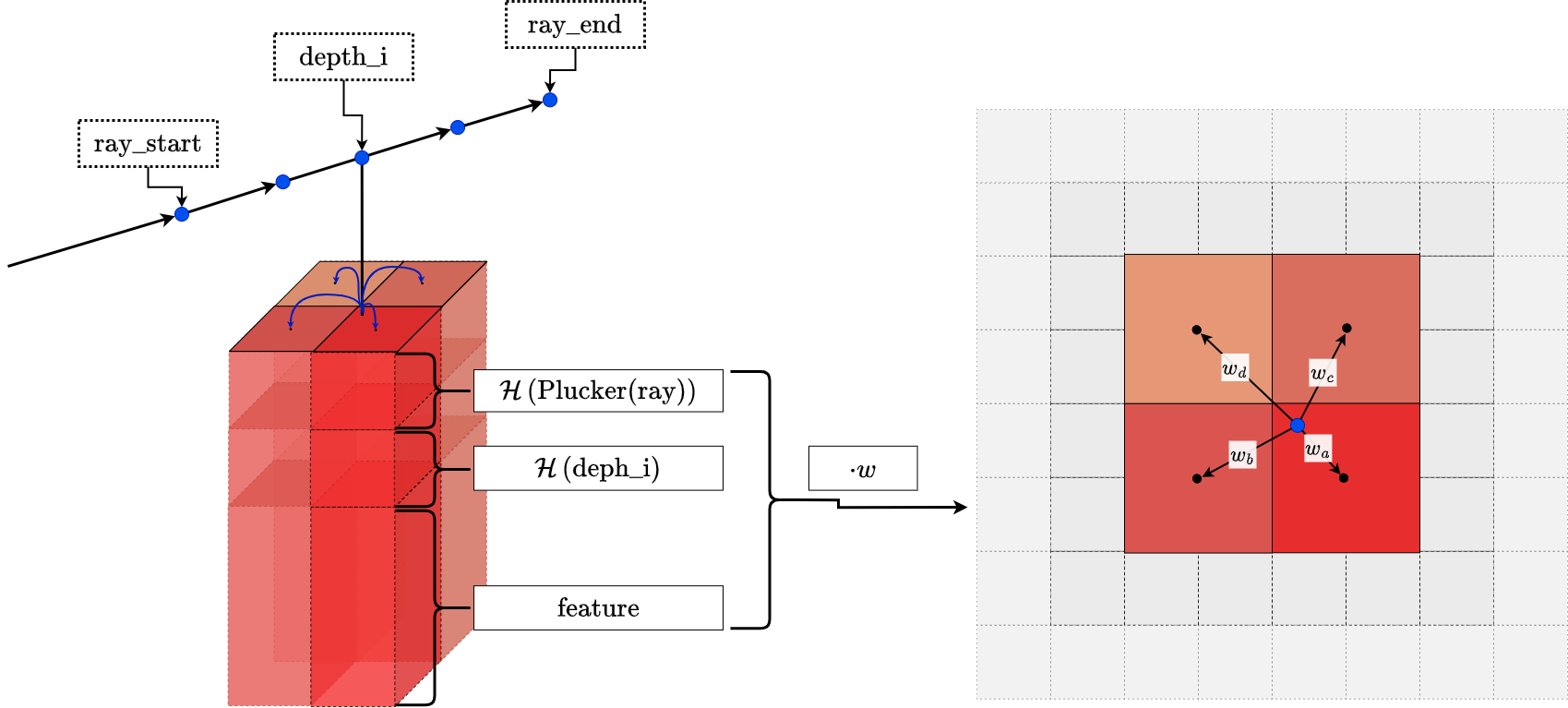}
    \caption{Each sampled point along the ray is enriched with the Harmonic Embedding of both the ray’s Plücker encoding and the sample depth, along with the pixel feature, before inverse bilinear splatting onto the planes.}
    \label{fig:inverse_bilinear_splatting}
\end{figure}

Along each ray, \(M\) points are uniformly sampled inside the cube. \textit{Each point acts as a probe that brings into space both the pixel feature it originates from and the geometric context of its camera (viewing direction and depth)}. For this reason, we associate each point \(q\) with a feature vector \( f_q \in \mathbb{R}^C \), obtained by concatenating the originating pixel feature \( f_{\text{pix}} \), the Harmonic Embedding \(\mathcal{H}(\cdot)\) of both the Plücker encoding of the ray direction and the depth of that point (Figure \ref{fig:inverse_bilinear_splatting}).

\begin{figure}[H]
    \centering
    \includegraphics[width=\columnwidth]{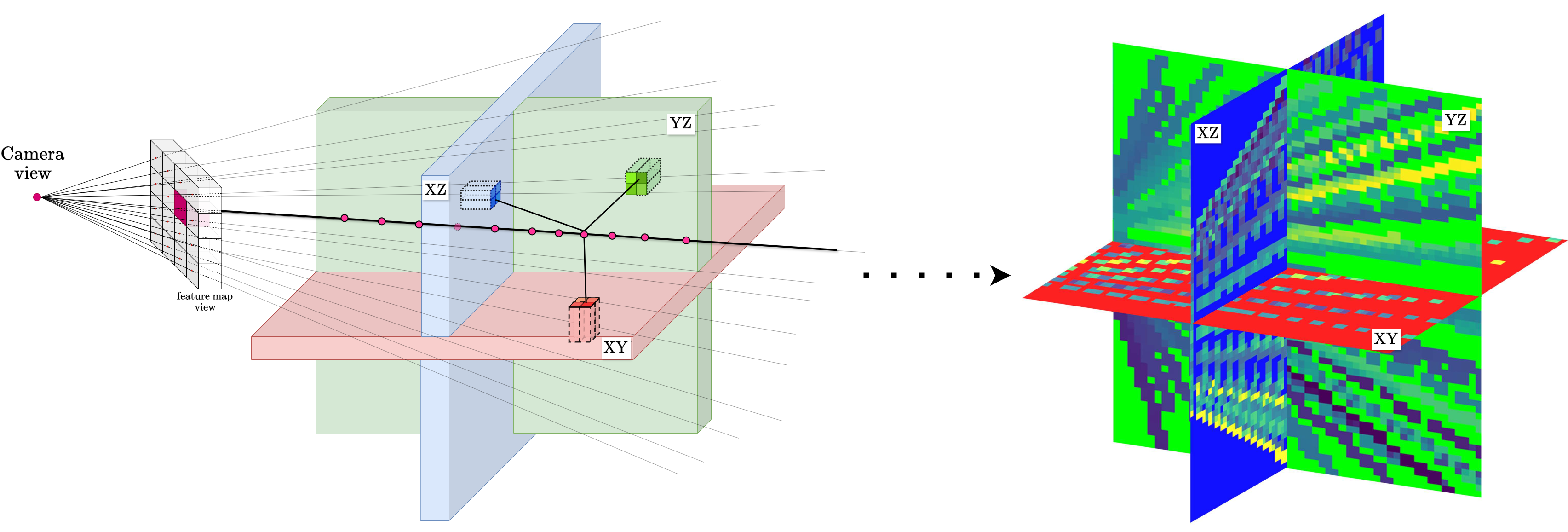}
    \caption{Sampling along rays and splatting of points onto three distinct planes per view, preserving separation between projections of different cameras.}
    \label{fig:plane_splatting}
\end{figure}

These enriched and contextualized vectors \(f_q\) are then projected onto three distinct latent planes \( \Pi \in \{XY, YZ, XZ\} \) via the projection function \( \pi(q) \in \mathbb{R}^2 \), indicating their position on each plane (Figure \ref{fig:plane_splatting}). Importantly, each view maintains \textit{its own set of three splatting planes}, which remain separate from those of other views.

Since the projection \( \pi(q) \) depends on the 3D spatial location and camera geometry, \textit{the splatting process inherently encodes the originating camera’s orientation}, thus providing spatially consistent feature encoding.

The contribution of each \( f_q \) is distributed separately on each plane through a process we call \textit{inverse bilinear splatting} (Figure \ref{fig:inverse_bilinear_splatting}), producing the splatted feature \( F_\Pi(p) \) at pixel \( p \) of plane \( \Pi \) as:
\begin{equation}
F_\Pi(p) = \frac{\sum_{q \in \mathcal{N}(p)} w(p, q) \cdot f_q}{\sum_{q \in \mathcal{N}(p)} w(p, q) + \varepsilon} \,,
\end{equation}
where \( w(p, q) \in [0,1] \) is the bilinear weight based on the distance between \( p \) and \( \pi(q) \), \(\varepsilon\) is a small stability term, and \(\mathcal{N}(p)\) denotes the set of points \(q\) contributing to the four nearest pixels around \(p\).

At the end of this stage, we obtain \(N\) sets of XY, YZ, and XZ planes—one set per camera/model—each containing that viewpoint’s current scene hypothesis.

Crucially, each model splats onto separate planes rather than shared ones. In the case where multiple views are splatted onto the same planes, features from different rays would overlap in the same pixels. Simple operations like summation or averaging would then cause detail loss and noise, which gets worse as the number of views \(N\) increases. To avoid this, we design a dedicated fusion stage that merges features without destructive blending.

\subsection{Fusion Stage}

In this stage, the \(N\) planes of each orientation—namely all XY, YZ and XZ planes—are fused separately to produce an initial visual consensus among the views.  
To avoid information loss, each pixel in the planes is aggregated using an attention-based strategy.

\begin{figure}[H]
    \centering
    \includegraphics[width=0.9\columnwidth]{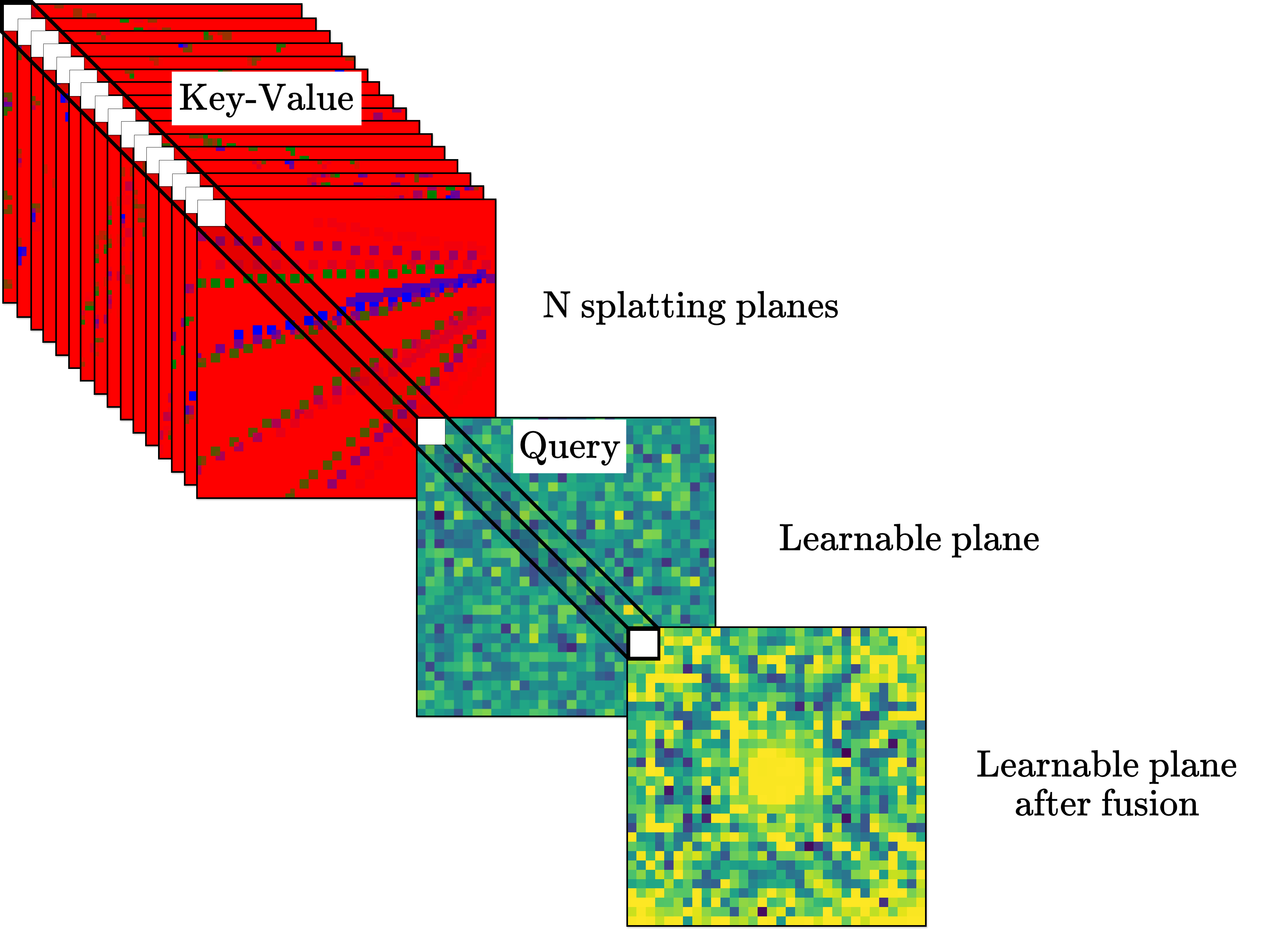}
    \caption{Pixel-wise cross-attention between the \(N\) splatting planes (top) and the learnable fusion plane (center), producing the final fused plane (bottom) that encodes a consistent visual and geometric synthesis across views.}
    \label{fig:fusion}
\end{figure}

Specifically, we introduce three additional learnable planes (one per orientation: XY, YZ and XZ), whose pixels are optimized during training.  
Each pixel in these fusion planes learns to aggregate the corresponding pixels from the \(N\) homologous planes—e.g., each pixel in the XY fusion plane aggregates pixels at the same location from all \(N\) XY planes—via cross-attention \cite{vaswani2017attention}, where the learnable pixel acts as the query and the corresponding pixels from the \(N\) planes serve as keys and values (Figure \ref{fig:fusion}).

Formally, let
$
A \in \mathbb{R}^{H \times W \times C}
$
be one of the learnable fusion planes (e.g., the XY plane) and
$
\{ B_i \}_{i=1}^N, \quad B_i \in \mathbb{R}^{H \times W \times C}
$
the corresponding \(N\) non-learnable splatting planes of the same orientation (e.g., all XY planes from the \(N\) views). We compute:
\begin{align}
q_{x,y} = A_{x,y} W_Q, \quad
k_{i,x,y} = B_{i,x,y} W_K, \quad
v_{i,x,y} = B_{i,x,y} W_V,
\end{align}
followed by cross-attention to update each pixel \((x,y)\) of \(A\):
\begin{align}
A'_{x,y} = \mathrm{CrossAttn}\big(
\underbrace{q_{x,y}}_{\text{query}}, \quad
\underbrace{\{ k_{i,x,y} \}_{i=1}^N}_{\text{keys}}, \quad
\underbrace{\{ v_{i,x,y} \}_{i=1}^N}_{\text{values}}
\big),
\end{align}
and a final projection:
\begin{align}
A''_{x,y} = A'_{x,y} W_O,
\end{align}

where \(W_Q, W_K, W_V \in \mathbb{R}^{C \times d}\) and \(W_O \in \mathbb{R}^{d \times C}\) are learnable projection matrices.

\subsection{Weaving Stage}

The three learnable fusion planes obtained from the previous stage represent a discrete consensus across views by attentively aggregating information per orientation (XY, YZ, XZ).  
However, some pixels may remain empty or incomplete when no views contribute at those locations. To build a continuous and robust latent representation, we perform a \textit{weaving} operation: an iterative, learned spatial refinement that integrates and smooths features across the planes.

This process begins with applying AdaLayerNorm (AdaLN) \citep{perez2018film, peebles2023scalable} independently to each plane, modulating features based on the current timestep to improve temporal consistency.

Next, features from all planes are jointly processed via self-attention, allowing spatial locations of each plane to interact and exchange information across the other orientations and then be refined by an MLP.

This weaving process is formally repeated as
\begin{align}
x' &= x + \mathrm{SelfAttention}(\mathrm{AdaLN}(x, t)) \\
x'' &= x' + \mathrm{MLP}(\mathrm{AdaLN}(x', t)) \,,
\end{align}

where \(x\) represents the spatial locations from all planes.

\begin{figure}[H]
    \centering
    \includegraphics[width=0.8\columnwidth]{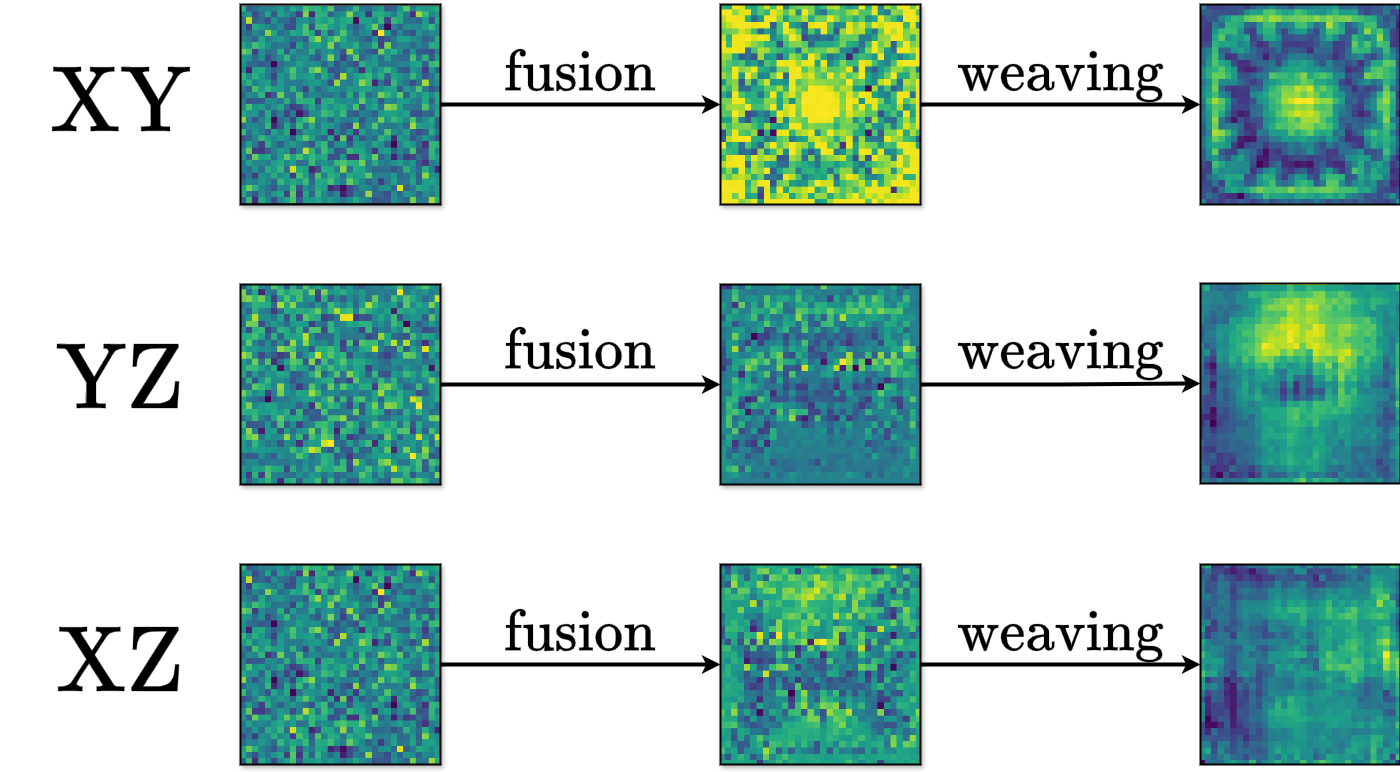}
    \caption{From raw learnable fusion planes to attentive fusion with splatted planes, followed by weaving to interpolate and consolidate the final representation.}
    \label{fig:weaving}
\end{figure}

The fusion and weaving stage are repeated multiple times within each decoder block, facilitating information integration and propagation between planes, improving spatial continuity and temporal coherence of the latent representation (Figure \ref{fig:weaving}).

\subsection{Latent Rendering Stage}

\begin{figure}[H]
    \centering
    \includegraphics[width=1.0\columnwidth]{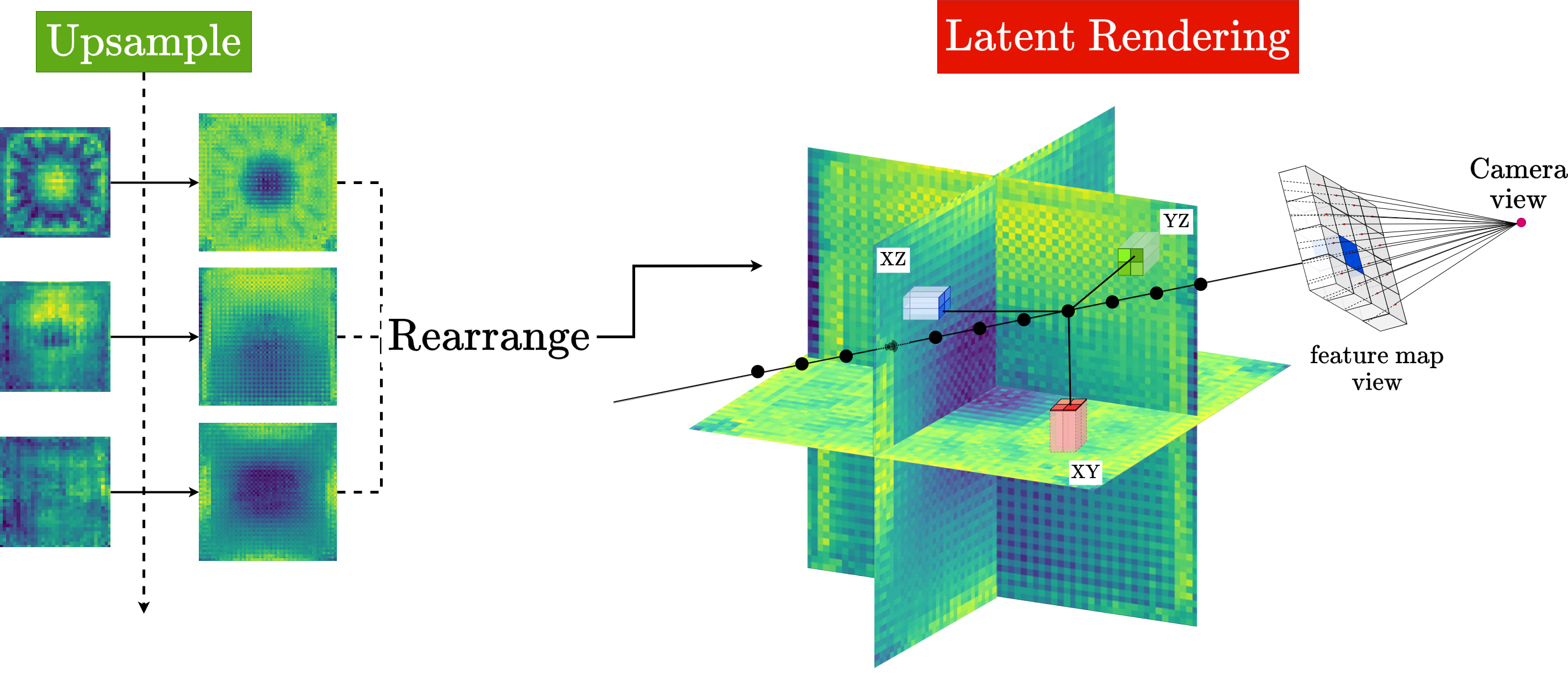}
    \caption{Features sampled along each ray define a global context used by a decoder to refine the original features. Corrections are aggregated via a weighted average.}
    \label{fig:latent_rendering}
\end{figure}

At the final decoder level, the processed XY, YZ and XZ planes are reorganized into a triplane used for the final view rendering (Figure \ref{fig:latent_rendering}).  
For each ray, we uniformly sample \(M\) points in space and interpolate features from the three planes. These features represent a \textit{point potentially observed by multiple views}, thus encoding a \textbf{shared global context} that enforces consistency across viewpoints.

For each source pixel \(p\), associated with a ray, let \(f_p\) be its original feature. We sample \(M\) points along the ray; for each point \(m=1,\dots,M\), we interpolate a global feature \(g_m \in \mathbb{R}^d\) from the three planes. Each pair \((f_p, g_m)\) is processed by a MLP decoder \(D\), inspired by EG3D \citep{chan2022efficient, hong2023lrm}

\[
\tilde{f}_m = D\big([f_p \,\|\, g_m]\big), \quad m=1,\dots,M,
\]
where \(\|\) denotes concatenation.

The corrected features \(\tilde{f}_m\) are then aggregated via a weighted average with normalized weights
\begin{align}
f_p^{\mathrm{corr}} = \sum_{m=1}^M w_m \tilde{f}_m,  \quad w_m = \frac{\exp(L(\tilde{f}_m))}{\sum_{j=1}^M \exp(L(\tilde{f}_j))}\,,
\end{align}
where \(L\) is a linear layer estimating each sample’s importance based on its feature.
The resulting \(f_p^{\mathrm{corr}}\) is the final, corrected pixel feature. The global features \(g_m\), coming from points potentially visible from multiple views, act as \textit{spatial guides} that steer the correction of \(f_p\) towards a representation coherent across views.

After \(\mathcal{T}\) diffusion steps, the final denoised latent features are decoded by \(\mathcal{D}\) to produce the RGB view.

\subsection{Final Loss}

The objective function combines the original diffusion loss from Zero1-to-3 (\ref{eq:loss_zero123}) with an additional \textit{Total Variation Loss (TVLoss)} \cite{shue20233d} applied to the learnable triplane fusion maps. This term acts as a spatial regularizer, penalizing abrupt local variations in the planes and encouraging smoothness and structural coherence in the learned features. The overall loss is defined as
\begin{equation}
\mathcal{L}_{\text{total}} = \mathcal{L}_{\text{diffusion}} + \lambda\cdot TV(XY, YZ, XZ) \,,
\end{equation}
where \(\lambda_{\text{TV}}\) balances the regularization strength against the main diffusion objective.




%% file: sec/4_experiment.tex
\section{Experiments}

\subsection{Implementation Details}

Each diffusion model is initialized with pre-trained weights from Zero-1-to-3 \cite{liu2023zero} and kept frozen to retain strong generalization ability; only the inter-model communication module is trained.

During both the \textit{per-view splatting} and \textit{latent rendering} stages, \(M=16\) points are uniformly sampled along each ray inside a cube centered at the origin with side length \(S=1.5\). The shared triplane processed at all decoder levels has resolution \(32 \times 32 \times 512\). Feature splatting planes per view start at resolution \(64 \times 64 \times C\), where \(C\) is the number of channels of the UNet at the current level, then downsampled to match the triplane resolution.

Both \textit{fusion} and \textit{weaving} stages use 8 heads with size 64 for each head. Both stages are applied multiple times per decoder block, with 3, 4, 6 and 8 iterations respectively in the bottleneck and subsequent levels of the decoder. The total variation regularization weight is \(\lambda=0.001\).

Training runs on 4 NVIDIA H100 GPUs (80GB each) with an initial learning rate of \(1 \times 10^{-5}\) and batch size 2 per GPU.





\subsection{Dataset and Evaluation Protocol}

We train on the LVIS subset of Objaverse~\cite{deitke2023objaverse}, which contains about 40K 3D models. For each object, we render 96 views at \(256 \times 256\) resolution from cameras positioned on six concentric rings at elevations \(-10^\circ\) to \(40^\circ\), with azimuth sampled uniformly in \([0^\circ, 360^\circ]\). During training, we randomly select N=16 views per object.

We evaluate on the Google Scanned Objects dataset~\cite{downs2022google}, consisting of over 1000 real scanned models, following EPiDiff’s~\cite{huang2024epidiff} protocol, that include two scenarios:
\begin{enumerate}
    \item \textbf{Fixed elevation:} elevation fixed at \(30^\circ\), azimuth uniform in \([0^\circ, 360^\circ]\).
    \item \textbf{Variable elevation:} elevation sampled between \(-10^\circ\) and \(40^\circ\), azimuth uniform.
\end{enumerate}

We report standard image quality metrics for evaluation of multi-view generation: PSNR, SSIM \cite{wang2004image} and LPIPS \cite{zhang2018unreasonable}.

To assess multi-view consistency, we reconstruct 3D meshes from generated views by extracting foreground masks using CarveKit and optimizing an Instant-NGP-based SDF~\cite{instant-nsr-pl} for 2000 steps ($\sim$1.5 min per object). We measure reconstruction quality with Chamfer Distance and Volumetric Intersection over Union.

\subsection{Baselines}

We compare against Zero-1-to-3 XL \cite{liu2023zero}, One-2-3-45 \cite{liu2023one}, Point-E \cite{nichol2022point}, Shap-E \cite{jun2023shap}, SyncDreamer \cite{liu2023syncdreamer} and EpiDiff \cite{huang2024epidiff}.
One-2-3-45 builds on Zero123 to generate multi-view images, from which it derives an SDF representation. Point-E and Shap-E are 3D generative models by OpenAI trained on large proprietary datasets. For Point-E, we use the \texttt{base1B} diffusion model that generates point clouds, which we convert to SDFs using official models. For Shap-E, we employ STF rendering mode, enabling joint prediction of SDF and colors to generate colored meshes directly.



\subsection{Discussion}

\begin{figure*}[t] 
    \centering
    \includegraphics[width=\textwidth]{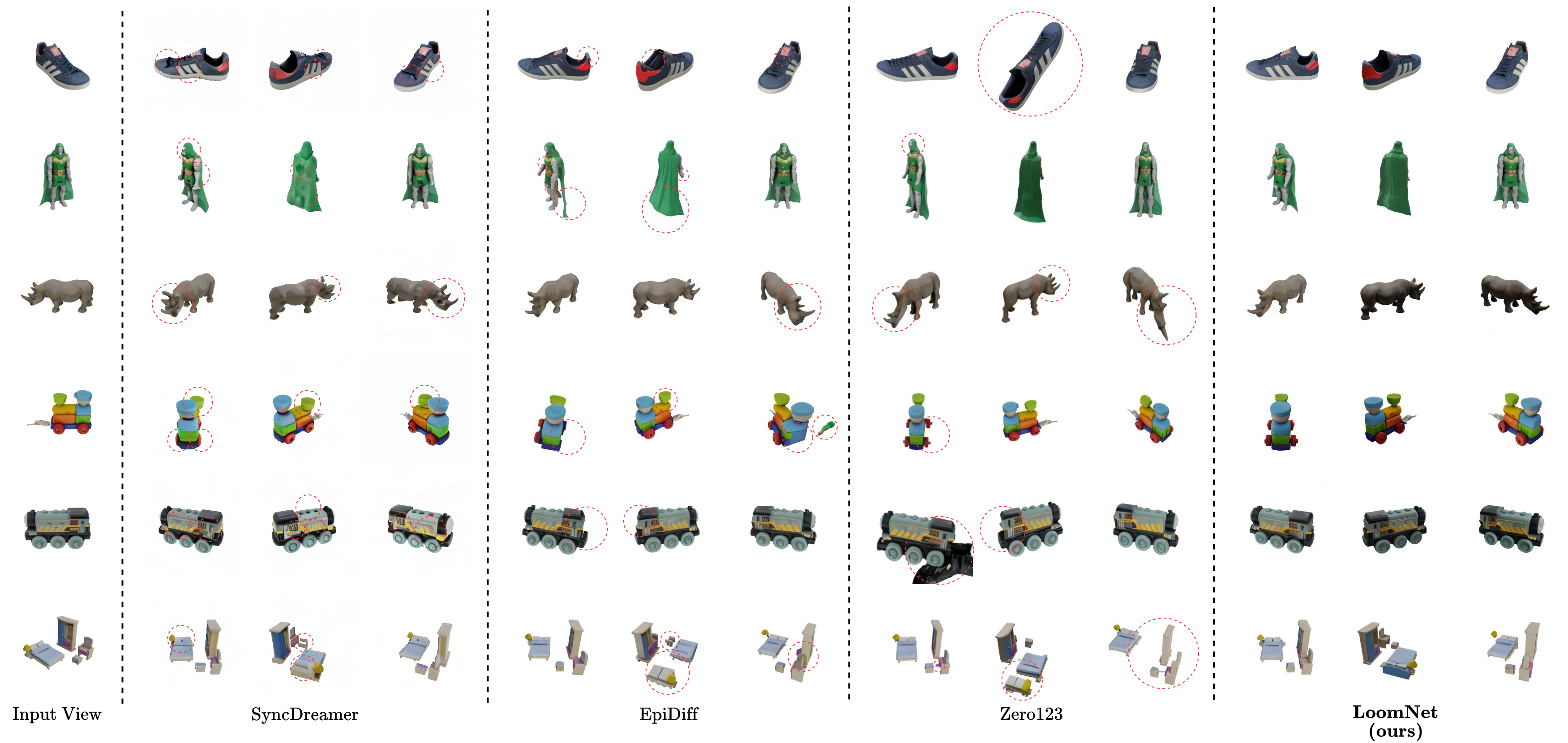}
    \caption{Qualitative comparison in generating novel views from a single image.}
    \label{fig:multiview_results}
\end{figure*}

\begin{figure*}[t] 
    \centering
    \includegraphics[width=0.9\textwidth]{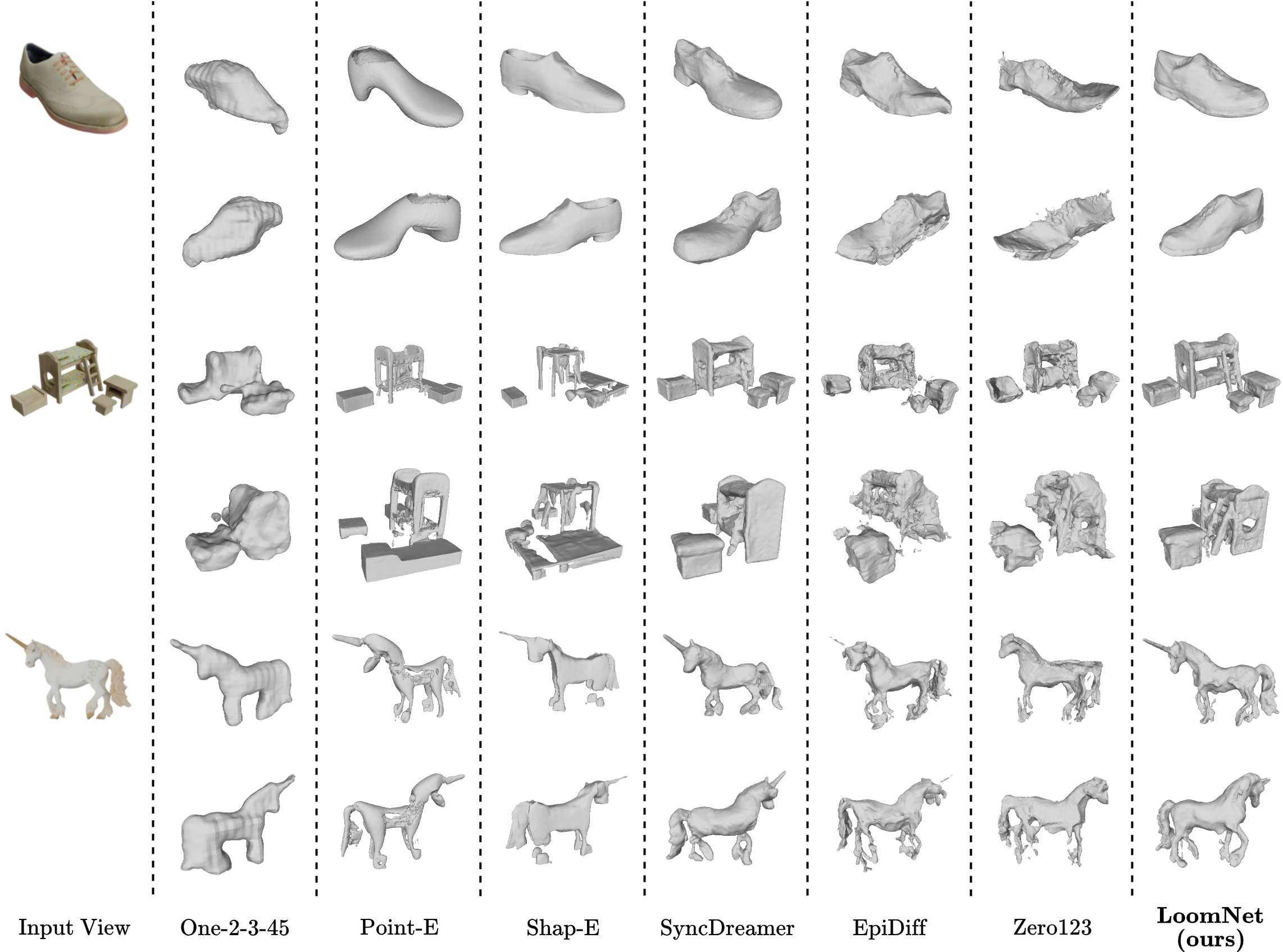}
    \caption{Qualitative comparison in surface reconstruction from single image.}
    \label{fig:mesh_results}
\end{figure*}

LoomNet achieves state-of-the-art results in both multi-view synthesis and 3D reconstruction, outperforming recent methods across all key metrics (Tables \ref{tab:multiview_comparison}, 
and \ref{tab:reconstruction_comparison}). Visually, generated views are more coherent and geometries more accurate (Figures \ref{fig:multiview_results}, \ref{fig:mesh_results}).

A key factor behind these results is the \textit{weaving stage}, which spatially integrates views by filling unobserved latent regions via structured feature interpolation. This yields a shared, continuous and consistent latent space even under complex camera distributions.

Notably, in the \textit{uniform setting} with irregular camera positions, LoomNet’s performance remains nearly unchanged compared to the fixed elevation (30°) case, demonstrating the weaving stage’s ability to generalize and effectively aggregate spatially distant information.

Moreover, our method is inference-efficient, running faster than many existing approaches while delivering superior quality.





\begin{table}[h]
\centering
\resizebox{\columnwidth}{!}{%
\begin{tabular}{lcccc}
\toprule
\textbf{Method} & \textbf{PSNR ↑} & \textbf{SSIM ↑} & \textbf{LPIPS ↓} & \textbf{Runtime ↓} \\
\midrule
\multicolumn{5}{l}{\textit{Fixed Elevation}} \\
\toprule
Zero123 \cite{liu2023zero} & 17.79 & 0.796 & 0.201 & \textbf{7s} \\
SyncDreamer \cite{liu2023syncdreamer} & 20.11 & 0.829 & 0.159 & 60s \\
EpiDiff \cite{huang2024epidiff} & 20.49 & 0.855 & 0.128 & 17s \\
\textbf{LoomNet (ours)} & \textbf{21.60} & \textbf{0.901} & \textbf{0.070} & 15s \\[1.2ex]
\multicolumn{5}{l}{\textit{Variable Elevation}} \\
\toprule
Zero123 \cite{liu2023zero} & 15.91 & 0.772 & 0.231 & \textbf{7s} \\
SyncDreamer \cite{liu2023syncdreamer} & 15.90 & 0.773 & 0.246 & 60s \\
EpiDiff \cite{huang2024epidiff} & 18.83 & 0.821 & 0.163 & 17s \\
\textbf{LoomNet (ours)} & \textbf{21.11} & \textbf{0.898} & \textbf{0.080} & 15s \\
\bottomrule
\end{tabular}
}
\caption{\textbf{Multi-View Synthesis Results.} Quantitative comparison under two scenarios on the GSO dataset: \textbf{Fixed Elevation} (30° setting) and \textbf{Variable Elevation} (uniform setting). Metrics reported are PSNR (↑), SSIM (↑), LPIPS (↓) and Runtime (↓).}
\label{tab:multiview_comparison}
\end{table}

\begin{table}[h]
\centering
\begin{tabular}{lcc}
\toprule
\textbf{Method} & \textbf{Chamfer Dist. ↓} & \textbf{Volume IoU ↑} \\
\midrule
One-2-3-45 \cite{liu2023one} & 0.0768 & 0.2936 \\
Point-E \cite{nichol2022point} & 0.0570 & 0.2027 \\
Shap-E \cite{jun2023shap} & 0.0689 & 0.2473 \\
Zero123 \cite{liu2023zero} & 0.0543 & 0.3358 \\
SyncDreamer \cite{liu2023syncdreamer} & 0.0496 & 0.4149 \\
EpiDiff \cite{huang2024epidiff} & 0.0429 & 0.4518 \\
\textbf{LoomNet} & \textbf{0.0260} & \textbf{0.5366} \\
\bottomrule
\end{tabular}
\caption{Quantitative comparison of surface reconstruction. We report Chamfer Distance (↓) and Volume IoU (↑) on the GSO dataset.}
\label{tab:reconstruction_comparison}
\end{table}

\subsection{Ablation Study}

We perform the ablation study under the fixed-elevation setting, varying one component at a time. The \textit{reference setup} (Table~\ref{tab:ablation}) uses latent rendering (LR) only at the final decoder layer, M=16 ray samples, positional encoding (PE) and cross-attention-based fusion of planes. Applying LR at all layers degrades quality. A simple fusion by averaging the planes reduces runtime and slightly degrades the other metrics. Halving samples marginally improves speed with minor quality loss. Removing PE causes the largest drop. Further details can be found in the supplementary materials.

\begin{table}[h]
\centering
\resizebox{\columnwidth}{!}{%
\begin{tabular}{lcccc}
\toprule
\textbf{Model Variant} & \textbf{PSNR ↑} & \textbf{SSIM ↑} & \textbf{LPIPS ↓} & \textbf{Runtime ↓} \\
\midrule
Reference Setup & \textbf{21.60} & \textbf{0.901} & \textbf{0.070} & 15s \\
LT at all decoder layers & 19.43 & 0.837 & 0.132 & 16.5s \\
Mean fusion of planes & 21.11 & 0.864 & 0.091 & \textbf{12.5s} \\
M=8 & 21.39 & 0.890 & 0.082 & 14s \\
w/o PE & 20.66 & 0.844 & 0.128 & 14s \\
\bottomrule
\end{tabular}%
}
\caption{Ablation results. The first row shows the reference configuration. Each subsequent row modifies one element from this setup, as indicated.}
\label{tab:ablation}
\end{table}


%% file: sec/5_limitation_and_conclusion.tex
\section{Limitations and Conclusion}

\subsection{Limitations}

Despite outperforming existing methods both quantitatively and qualitatively, LoomNet has limitations. In particular, it is less effective than methods like SyncDreamer at propagating errors consistently across views. While SyncDreamer may generate less accurate and quality images, its shared latent space leads to higher inter-view consistency, often resulting in smoother—though sometimes less semantically accurate—meshes.

Moreover, the 3D reconstruction process happens in two separate stages: view generation and reconstruction. A unified pipeline could further improve consistency and also efficiency.

\subsection{Conclusion}

We presented \textbf{LoomNet}, a multi-view diffusion model designed to improve cross-view consistency by enabling communication between independently generated viewpoints. Our approach leverages per-view splatting to project scene hypotheses onto latent planes, which are then fused and refined through a weaving stage into a unified latent scene embedding. This shared representation enables consistent latent rendering across views. Experimental results confirm that LoomNet achieves superior view coherence, resulting in significant improvements in 3D reconstruction quality, both in terms of mesh fidelity and mesh quality, while maintaining fast inference times.